\documentclass[pdflatex,sn-mathphys-num]{sn-jnl}% Math and Physical Sciences Numbered Reference Style
\usepackage{graphicx}%
\usepackage{multirow}%
\usepackage{amsmath,amssymb,amsfonts}%
\usepackage{amsthm}%
\usepackage{mathrsfs}%
\usepackage[title]{appendix}%
\usepackage{xcolor}%
\usepackage{textcomp}%
\usepackage{manyfoot}%
\usepackage{booktabs}%
\usepackage{algorithm}%
\usepackage{algorithmicx}%
\usepackage{algpseudocode}%
\usepackage{listings}%
\usepackage{multirow}

\usepackage{url}
\usepackage{adjustbox} 
\usepackage{enumitem}
\usepackage{multirow}
\usepackage{newfloat}
\usepackage{hyperref}

\theoremstyle{thmstyleone}%
\theoremstyle{thmstyletwo}%

\theoremstyle{thmstylethree}%

\begin{document}

\title[KGCaRe: Explainable Complex Conditional Question
Answering using Automatic Knowledge Graph Construction
and Context Retrieval with LLMs]{KGCaRe: Explainable Complex Conditional Question
Answering using Automatic Knowledge Graph Construction
and Context Retrieval with LLMs}

%%=============================================================%%
%% GivenName	-> \fnm{Joergen W.}
%% Particle	-> \spfx{van der} -> surname prefix
%% FamilyName	-> \sur{Ploeg}
%% Suffix	-> \sfx{IV}
%% \author*[1,2]{\fnm{Joergen W.} \spfx{van der} \sur{Ploeg} 
%%  \sfx{IV}}\email{iauthor@gmail.com}
%%=============================================================%%

\author*[1]{\fnm{Ghanshyam} \sur{Verma}} \email{ghanshyam.verma@universityofgalway.ie}

\author[1]{\fnm{Simanta} \sur{Sarkar}}\email{simanta.sarkar@universityofgalway.ie}

\author[1]{\fnm{Devishree} \sur{Pillai}} \email{devishree.pillai23@gmail.com}

\author[2]{\fnm{Hotaka} \sur{Shiokawa}}\email{hotaka.shiokawa@fmr.com}
%\equalcont{These authors contributed equally to this work.}

\author[2]{\fnm{Yourong} \sur{Xu}}\email{yourong.xu@fmr.com}

\author[3]{\fnm{Fiona} \sur{Veazey}}\email{fiona.veazey@fmr.com}

\author[3]{\fnm{Peter} \sur{Hubbert}}\email{peter.hubbert@fmr.com}

\author[2]{\fnm{Hui} \sur{Su}}\email{Hui.Su@fmr.com}

\author[1]{\fnm{Paul} \sur{Buitelaar}}\email{paul.buitelaar@universityofgalway.ie}

\affil*[1]{\orgdiv{Insight Research Ireland Centre for Data Analytics, Data Science Institute}, \orgname{University of Galway}, \orgaddress{\street{IDA Business Park}, \city{Galway}, \postcode{H91 AEX4}, \state{Co. Galway}, \country{Ireland}}}

\affil[2]{\orgname{Fidelity Investments}, \orgaddress{\city{Boston}, \state{Massachusetts}, \country{USA}}}

\affil[3]{\orgname{Fidelity Investments}, \orgaddress{\city{Dublin}, \state{Co. Dublin}, \country{Ireland}}}

%\affil[3]{\orgdiv{Fidelity Investments}, \orgname{Organization}, \orgaddress{\street{Street}, \city{City}, \postcode{610101}, \state{State}, \country{Country}}}

%%==================================%%
%% Sample for unstructured abstract %%
%%==================================%%

\abstract{Answering complex conditional questions using Large Language Models (LLMs) and Retrieval-Augmented Generation (RAG) remains a challenge, particularly in domain-specific contexts where general-purpose LLMs and RAG tend to underperform. We hypothesize that augmenting RAG with unstructured and structured knowledge, extracted from both documents and knowledge graphs (KGs), can improve reasoning and answer accuracy for such tasks.

To test this, we propose KGCaRe, a hybrid approach that combines neural retrieval with symbolic reasoning over LLM-generated KGs. KGCaRe constructs a KG from documents using a multi-prompt extraction strategy and stores it in a graph database. Simultaneously, the documents are embedded into a vector store to enable neural retrieval. KGCaRe performs innovative iterative graph traversal guided by the LLM to extract relevant triples, prune irrelevant information, and uses additional clue entities to traverse the graph again if the initial traversal does not provide satisfactory context to generate the answer. The relevant triples extracted from the KG in path form, along with semantically retrieved text passages, are then fed into custom KGCaRe prompts to generate answers to the complex conditional questions with explanations.

We evaluate KGCaRe on two complex conditional QA datasets. Our results on these datasets show that KGCaRe consistently outperforms existing baselines, including Vanilla LLM, Code Prompt, Text Prompt, Think-on-Graph, Vanilla RAG, and HybridContextQA, across multiple LLMs such as Mistral, Mixtral, GPT-3.5, and GPT-4o. We publicly release the software pipeline that we developed to implement the proposed KGCaRe approach.}

%%================================%%
%% Sample for structured abstract %%
%%================================%%


%%\pacs[JEL Classification]{D8, H51}

%%\pacs[MSC Classification]{35A01, 65L10, 65L12, 65L20, 65L70}

\maketitle

\section{Introduction}
Answering domain-specific conditional questions using LLMs presents a significant challenge. Although LLMs perform well on general-purpose QA tasks, their effectiveness diminishes considerably when the input involves conditional logic or requires reasoning across domain-specific content \citep{pan2024unifying, 10.5555/3618408.3619049}. One of the reasons behind this performance drop is the absence of fine-tuning on specialized corpora. Additionally, when presented with insufficient context, LLMs are prone to hallucinate—generating answers that appear plausible but lack grounding in the source material \cite{zhang2023siren}. Conditional questions are particularly sensitive to missing or implicit constraints; the LLM must be aware of both the question and the conditions under which an answer is valid. These challenges point to the need for enhanced strategies that go beyond prompting or fine-tuning.

A promising direction is to incorporate external knowledge sources such as knowledge graphs (KGs), which organize information in semantically structured, explainable formats \cite{nickel2015review, wang2017knowledge}. KGs can complement LLMs by providing factual precision and enabling symbolic reasoning \cite{10.1145/3686806}. However, extracting high-quality triples from raw domain-specific documents can be error-prone and incomplete, limiting the usefulness of the KG. On the other hand, semantic vector retrieval from the same documents offers a flexible way to supplement the LLM’s understanding with textual evidence. This motivates our hypothesis: a hybrid context composed of both symbolic and neural representations could improve performance on complex conditional QA tasks.

To validate this, we propose a new approach called KGCaRe (\textbf{K}nowledge \textbf{G}raph \textbf{C}ontext \textbf{a}ware \textbf{Re}asoning for complex Question Answering). It combines symbolic retrieval from a LLM-generated KG with vector-based semantic search from the same document corpus. The KG is constructed using a multi-stage prompt strategy and stored in a Neo4j database \cite{vukotic2015neo4j,monteiro2023experimental}, while the vector index is built using embeddings stored in Facebook AI Similarity
Search (FAISS) \cite{douze2024faiss} supported vector storage. During inference, KGCaRe performs iterative graph traversal over the KG to identify relevant triples and combine them with semantically retrieved passages. This fused context is then used to generate answers that are both accurate and explainable.

We evaluate KGCaRe using two publicly available QA datasets: ConditionalQA \cite{sun-etal-2022-conditionalqa} and HotpotQA \cite{yang2018hotpotqa}. ConditionalQA contains complex, multi-hop, conditional questions derived from UK policy documents. HotpotQA contains complex multi-hop questions derived from Wikipedia articles. We also conduct an ablation study to evaluate the performance of the proposed and existing approaches on specific types of questions. 

The key contributions of this work are:
\begin{itemize}
\item We design and implement an end-to-end pipeline for automatic KG construction from long documents with complex structure, using a multi-prompt LLM-based triple extraction approach.
\item We introduce KGCaRe, a hybrid approach that leverages both KG and vector context for answering complex conditional questions.
\item We empirically show that KGCaRe outperforms baseline methods such as Vanilla LLM, Code Prompt, Text Prompt, Think-on-Graph, Vanilla RAG, and HybridContextQA across multiple LLMs.
\item We publicly release the end-to-end software pipeline code and the associated prompts used to implement our approach\footnote{\url{https://github.com/GhanshyamVerma/KGCaRe}}, to support reproducibility and further research.
\end{itemize}

The rest of the paper is structured as follows. In Section \ref{section:RelatedWork}, we describe related work. Section \ref{Label_Data} describes the ConditionalQA and HotpotQA datasets. In Section \ref{Label_Approch}, we explain our proposed approach. Section \ref{Label_Exp_Design} describes the experimental design that we used. In Section \ref{sec_results}, we discuss and compare results in detail. Finally, we conclude in Section \ref{sec_conclusions}.

\section{Related Work}  \label{section:RelatedWork}
Despite impressive gains in general-domain question answering, LLMs continue to struggle with tasks requiring multi-hop, conditional reasoning over long and complex documents. In this context, the ConditionalQA dataset introduced by Sun et al. \cite{sun-etal-2022-conditionalqa} stands out for its focus on realistic, policy-based QA that includes multiple reasoning steps and conditional constraints.

Puerto et al. proposed a code prompting strategy \cite{puerto2024code} that converts natural language queries and context from documents into code, which is then used as input to the LLM prompts for answer generation. While this approach provides the query and context in a more machine-understandable format, it lacks explicit incorporation of structured knowledge, such as knowledge from KGs.

Combining KGs with LLMs is a growing area of research. Prior works have shown that knowledge graphs can enhance factual grounding, improve faithfulness, and support reasoning in domains like law and medicine \cite{pan2024unifying}. For instance, MindMap \cite{wen2023mindmap} integrates biomedical triples from EMCKG into LLM prompts and shows gains on medical QA benchmarks.

Think-on-Graph \cite{sun2023think} and Think-on-Graph 2.0 \cite{ma2024think} perform iterative beam search with the help of LLMs to extract context from the KG and use that context to generate the answers. Luo et al. proposed a method called Reasoning on Graphs (RoG) that synergises LLMs with KGs so that faithful and interpretable reasoning can be performed \cite{luo2023reasoning}. These approaches can uncover multi-hop paths, but they often depend on dense graph connectivity. In contrast, our approach works even when the KG is sparse by supplementing it with semantically retrieved passages.

Retrieval-Augmented Generation (RAG) frameworks have also been explored as a solution for grounding LLMs in external documents \cite{gao2023retrieval}. However, most RAG implementations rely on simple keyword or similarity search, which may fail to retrieve critical conditions or logical dependencies \cite{sawarkar2024blended}. 

HybridContextQA \cite{verma2024hybridcontextqa} is an existing RAG-based hybrid approach that uses context from both documents and KGs to generate answers to complex and conditional questions. While HybridContextQA has shown promising results on the ConditionalQA dataset, it performs simple keyword-based search to extract context from KGs.

Tree-of-Traversal \cite{markowitz2024tree} is a zero-shot approach that enables augmentation of LLMs with one or more KGs to perform tree search over KGs. This approach significantly improves performance on KG question answering tasks; however, it does not consider context extraction from documents for answer generation.

To handle these issues, our KGCaRe approach combines symbolic graph traversal and semantic retrieval into a unified pipeline for better answer generation of complex and conditional questions.

\section{Datasets} \label{Label_Data}
We evaluate our approach on two publicly available benchmark complex QA datasets. The first dataset is ConditionalQA \cite{sun-etal-2022-conditionalqa}, which has been specifically designed to test a model’s capability in multi-hop reasoning and conditional answer generation. The dataset is constructed from UK public policy documents and reflects realistic scenarios where understanding legal and procedural nuances is essential for answering the questions correctly.

The ConditionalQA dataset includes four answer categories: yes/no answers, span-based answers that involve extracting text from the document, conditional answers which are only valid when certain criteria are met, and not-answerable cases where the document does not contain sufficient information to produce a valid response. One of the key challenges in ConditionalQA is that conditions relevant to the answer are often not explicitly stated in the question but are buried within various sections of the document. As a result, a model must perform multi-hop reasoning to locate and assemble these conditions into a coherent justification for its answer. 

The second QA dataset is HotpotQA \cite{yang2018hotpotqa}, a publicly available large-scale complex QA dataset that requires multi-hop reasoning to answer the questions. This dataset mainly contains two types of multi-hop questions: yes/no type and span type. In this dataset, there is no specific category of conditional-type questions that require explicit conditions to be generated along with the answer. However, the questions require implicit conditions to be checked in order to generate accurate answers, or involve comparison between two or more entities based on shared properties \cite{yang2018hotpotqa}. Please refer to Appendix \ref{sec:Data_Appendix} for further details on the datasets.

% \textbf{The second QA dataset is created by domain experts using proprietary data from a large multinational financial services company. Similar to ConditionalQA, this dataset also contains long legal documents, and the evidence required to answer a question can be scattered across multiple sections, tables, and complex document structures. We cannot release the created QA dataset or the associated documents, as they contain sensitive information about the company and its customers.}

\section{Proposed Approach} \label{Label_Approch}
KGCaRe introduces a comprehensive and modular architecture designed to support explainable reasoning across both unstructured and structured sources of information. The core of the KGCaRe approach lies in its hybrid retriever that seamlessly integrates symbolic reasoning over a knowledge graph with semantic similarity-based vector retrieval. Below, we detail each component of the proposed KGCaRe approach.

\subsection{Multi-Prompt Triple Extraction and Knowledge Graph Construction}
To enable precise and context-rich knowledge representation, KGCaRe employs a multi-stage, prompt-driven approach for KG construction using LLMs. This process is designed to extract highly accurate and logically connected (subject, predicate, object) triples from raw document text, using a pipeline of progressive prompt stages. These prompt stages are as follows:

\begin{itemize}
  \item \textbf{Stage 1: Contextual Understanding and Initial Entity-Relation Extraction} \\ The first stage uses a prompt (MLT\_PROMPT\_1) that instructs the LLM to analyze the input text deeply, identify key entities, and extract all possible relations. This prompt emphasizes understanding temporal, causal, procedural, and conditional constructs within the document. It encourages the model to extract granular, context-aware triples by first summarizing the purpose and conditions within the text and then listing entity-relationship pairs. For example: \\
  (`Applicant', `must submit', `form') \\
  (`form', `must be submitted within', `30 days') \\

  \item \textbf{Stage 2: Conditional and Alternative Scenario Expansion} \\
  Next, a second prompt (MLT\_PROMPT\_2) is used to enhance the previously extracted content by focusing on conditional logic, exceptions, and alternatives within the text. This step is crucial for capturing real-world nuances such as regulatory conditions, procedural dependencies, and branching decision logic. The prompt instructs the LLM to explicitly recognize if-then structures, alternatives, and exceptions, and to augment the triple set accordingly. For instance: \\
  Condition: If consent is not given, permission must be obtained from the court. \\
  Triples: (`Consent', `not given', `get permission from court')

  \item \textbf{Stage 3: Refinement and Logical Graph Expansion} \\
  The third prompt (MLT\_PROMPT\_3) refines the extracted triples by validating their completeness, enhancing specificity, and interlinking them for better logical consistency. It pushes the LLM to derive new triples from the previously extracted set using transitive or implied logic. This step ensures that the final graph includes both explicitly stated and inferable relationships. For example: \\
  Given: \\
(`Guardian', `needs consent from', `Parents') \\
(`Parents', `can delegate consent to', `Court of Protection') \\
Inferred: \\
(`Guardian', `can delegate consent to', `Court of Protection')

\item \textbf{Stage 4: Output Normalization and Graph Ingestion} \\
The final output from these prompts is formatted as a set of triples that preserve the semantics of the source text. These triples are then standardized and ingested into a Neo4j graph database, enabling scalable storage and advanced traversal logic during the retrieval phase.
This multi-prompt approach not only improves the quality and structure of extracted knowledge but also allows the KG to represent intricate dependencies and nuanced regulatory logic that single-pass extraction pipelines may miss (see Appendix \ref{sec:KG_Appendix}).

\end{itemize}

\subsection{Semantic Vector Indexing}
In parallel with KG construction, documents are embedded into high-dimensional vector representations using an LLM-powered embedder. These embeddings are stored in a FAISS index \cite{douze2024faiss}, allowing fast approximate nearest-neighbor search. This vector index captures the global semantic context of the documents and is used to retrieve relevant passages based on the similarity to the user’s query.

% \subsection{Knowledge Graph Traversal}
% The core innovation in KGCaRe lies in the proposed algorithm (see Algorithm 1), which implements a symbolic, LLM-guided traversal over the constructed KG. Given a query, the retriever first extracts candidate topic entities using keyword or entity detection. These entities are used to initialize the traversal process.  \\

\subsection{Knowledge Graph Traversal}

The central component of KGCaRe is its symbolic, LLM-guided KG traversal mechanism, formally outlined in Algorithm~\ref{alg:kg-care-qa}. This traversal process enables KGCaRe to iteratively explore and reason over the constructed KG to extract relevant contextual information for complex question answering.

Given an input question $q$, the pipeline begins with a preprocessing step where topic entities ($E_{topic}$) are extracted using an LLM. These topic entities act as anchors for initiating graph traversal. KGCaRe then initializes multiple memory modules, including the triple memory $M$, clue entity memory $E_{clue}$, clue triple memory $M_{clue}$, traversal path memory $M_{path\_traversed}$, and $visited\_entities$. 

Traversal proceeds in a depth-bounded iterative manner up to a maximum depth $D$. At each level, KGCaRe searches for triples within the KG. For the initial depth ($d=1$), the retrieval is based on partial matching to allow broader exploration, including subword matches. For subsequent depths ($d > 1$), exact matching is used to ensure more focused reasoning. In parallel, KGCaRe also searches for triples linked to clue entities ($E_{clue}$) using partial matching, enabling the discovery of supporting evidence not captured by topic entities alone. The use of the clue entity in our KGCaRe approach is inspired by the work of \cite{markowitz2024tree}, who proposed the Tree-of-Traversal approach.

\begin{algorithm}[H]
\small
\caption{Knowledge Graph Traversal in KGCaRe}
\label{alg:kg-care-qa}
\textbf{Input}: $q$ (input question), $D$ (maximum depth), $LLM$ (language model)\\
\textbf{Output}: $triples_{pruned}$, Memory ($M$)
\begin{algorithmic}[1]
\State Initialize $E_{topic}$ from $q$, memory $M$, clue entities $E_{clue}$, clue memory $M_{clue}$, path memory $M_{path\_traversed}$, and $visited\_entities$
\For{$depth = 1$ to $D$}
    \ForAll{$e_{topic_i} \in E_{topic}$}
        \If{$depth == 1$}
            \State Find partial matches for $e_{topic_i}$ with $visited\_entities$
        \Else
            \State Find exact matches for $e_{topic_i}$ with $visited\_entities$
        \EndIf
    \EndFor
    \ForAll{$e_{clue_i} \in E_{clue}$}
        \State Find partial matches for $e_{clue_i}$ with $visited\_entities$
    \EndFor
    \State $triples_{pruned} \gets$ PruneTriples($LLM$, $q$, extracted triples)
    \State Save candidates for next round using $triples_{pruned}$, $i$, $M$, $M_{path\_traversed}$
    \State Update $visited\_entities$, $M$, and $M_{path\_traversed}$
    \If{$triples_{pruned} \neq \emptyset$}
        \State Perform Reasoning($LLM$, $q$, $triples_{pruned}$, $M_{clue}$)
        \If{knowledge is sufficient}
            % \State Return triples wer($q$, $M$)
            \State \Return $triples_{pruned}$, Memory ($M$)
        \Else
            \State Update $E_{clue}$ and $M_{clue}$ with new clues
            \If{$E_{clue} \neq \emptyset$ or $E_{topic} \neq \emptyset$}
                \State \textbf{continue}
            \Else
                \State \textbf{break}
            \EndIf
        \EndIf
    \Else
        \State Perform Reasoning($LLM$, $q$)
        \State Update $E_{clue}$ and $M_{clue}$ with clues found
        \If{$E_{clue} \neq \emptyset$ or $E_{topic} \neq \emptyset$}
            \State \textbf{continue}
        \Else
            \State \textbf{break}
        \EndIf
    \EndIf
\EndFor
\State \Return $triples_{pruned}$, Memory ($M$)
\end{algorithmic}
%\vspace{-.2em}
\end{algorithm}

The retrieved triples are then passed through an LLM-powered pruning function:
\[
triples_{pruned} \gets \text{PruneTriples}(LLM, q, \text{triples})
\]
This step filters out irrelevant or redundant triples, retaining only those deemed contextually useful for answering the input question. The pruned triples are saved to the memory $M$ and are also used to update the visited entity set and traversal path memory.

Next, an LLM reasoning prompt is used to evaluate whether the accumulated context is sufficient to answer $q$. If so, KGCaRe generates the final answer and halts traversal. If not, the LLM is prompted to extract new clue entities that may assist in further traversal. These new entities are appended to $E_{clue}$, and the process continues to the next depth level. If no clue or topic entities remain, or no relevant triples are found, KGCaRe triggers a fallback mode where the LLM performs reasoning directly over the question to extract additional clues for KG traversal.

If traversal reaches the maximum depth $D$ or sufficient context is retrieved from the KG to generate the answer, it returns the KG context. KGCaRe invokes a final mechanism, where KGCaRe uses an LLM with a custom prompt to generate the answer using all accumulated information from memory $M$ (context from KG) and the context retrieved from the vector store.

Throughout the process, KGCaRe maintains an explicit reasoning trace, including the triples retrieved, pruned, and traversed, as well as the order in which entities were visited. This traceability allows KGCaRe to generate not only an answer but also an explanation of the logical steps and conditions leading to that answer, ensuring interpretability and transparency.

By dynamically combining symbolic graph traversal with LLM-guided clue generation and pruning, KGCaRe enables robust, multi-hop reasoning over sparse or incomplete KGs in complex conditional question answering scenarios.

\subsection{Combined Retrieval and Answer Generation}

Once relevant information has been gathered from both the KG traversal and the semantic vector store, KGCaRe merges these two context sources into a unified prompt for final answer generation. This fusion stage ensures that the LLM benefits from the structured precision of symbolic reasoning and the semantic richness of neural retrieval.

The graph-based retriever contributes a curated set of triples and their associated traversal paths, representing the logical structure of the reasoning process. Simultaneously, the semantic retriever returns top-ranked passages from the FAISS vector index, capturing global textual context relevant to the question. These passages may include supporting explanations, alternate phrasings, or implicit conditions that are not explicitly modeled in the KG.

Both the symbolic and neural contexts are formatted into a custom structured prompt that instructs the LLM to synthesize a coherent and accurate answer. The prompt explicitly asks the model to incorporate evidence from both sources and, where applicable, to identify and state any conditions under which the answer holds. This is particularly important for conditional questions, where part of the answer often depends on latent assumptions or constraints embedded in the retrieved context.

By combining symbolic and neural retrieval, KGCaRe improves not only answer accuracy but also explanation quality. This integrated reasoning strategy is essential for complex multi-hop questions that require connecting disparate facts and resolving conditional logic, especially in domains such as public policy, law, or regulations where answer faithfulness and traceability are critical.

\section{Experimental Design} \label{Label_Exp_Design}

To evaluate the effectiveness of KGCaRe, we design a series of experiments to compare against several baselines and existing state-of-the-art methods. The objective of these experiments is to examine how well KGCaRe performs in answering complex, conditional questions compared to existing KG-based, prompt-based, and RAG-based approaches.

We conduct comparative evaluations against two baseline prompting techniques: \textit{Text Prompt} and \textit{Code Prompt}, as introduced by Puerto et al. \cite{puerto2024code}. These approaches operate solely on raw document context without incorporating any structured knowledge. In parallel, we benchmark our model against \textit{Think-on-Graph} \cite{sun2023think}, which utilizes a KG to guide the LLM through a structured reasoning process. This comparison is crucial to assess the improvements brought by our hybrid design over purely symbolic KG traversal methods. 

We also perform comparative evaluations against Vanilla LLM, where the LLM is prompted directly using the raw question without any external context, and Vanilla RAG, where a standard RAG pipeline retrieves passages from the documents and feeds them into the LLM without incorporating any KG-based reasoning. These experiments are conducted using four LLMs: Mistral \cite{jiang2023mistral7b}, Mixtral \cite{jiang2024mixtralexperts}, GPT-3.5, and GPT-4o. Furthermore, we evaluate our method against an existing hybrid approach called HybridContextQA \cite{verma2024hybridcontextqa}.

We performed the evaluation using two datasets: ConditionalQA and HotpotQA. For both datasets, the development set is used as the evaluation benchmark for all comparative experiments, as the test set is not publicly available. ConditionalQA consists of 2,338 training QA pairs and 271 development QA pairs, each associated with UK policy documents. HotpotQA consists of 90,564 QA pairs in the training set and 500 QA pairs in the development set. Please refer to Appendix \ref{sec:Data_Appendix} for further details on the datasets. We use some QA examples from the training set for few-shot prompting to provide the LLMs with in-context examples.

For all experiments, we test KGCaRe and baseline models using four different LLMs: GPT-3.5, GPT-4o, Mistral, and Mixtral. Please refer to Appendix \ref{sec:Implementation_Appendix} for implementation details and Appendix \ref{sec:Experimental_Appendix} for experimental environment details. This diverse set of LLMs allows us to evaluate the generalizability of our approach across both proprietary and open-source LLMs, and to observe how model scale and architecture influence the effectiveness of hybrid retrieval strategies. Similar to \cite{verma2024hybridcontextqa}, we use the exact match based F1 score as an evaluation metric. Please refer to Appendix \ref{sec:Evaluation_Appendix} for further details on the evaluation metric.

Through these experiments, we aim to provide a comprehensive analysis of KGCaRe’s strengths and limitations in the context of complex and conditional QA task, and to demonstrate its consistent advantages over both RAG and KG-only baselines.

\section{Results} \label{sec_results}

We evaluate KGCaRe using two datasets across four different LLMs: Mistral, Mixtral, GPT-3.5, and GPT-4o. The ConditionalQA dataset contains 271 QA pairs spanning Yes/No, span-based, and conditional answer types (see Table~\ref{table:all_results}). The performance of the proposed KGCaRe is compared against several baselines, including Vanilla LLM, Code Prompt, Text Prompt, Think-on-Graph, Vanilla RAG, and HybridContextQA.

\begin{table*}[b]
%\vspace{-0.5em}
\centering
\small
\caption{Results of different prompting and context integration approaches for various LLMs on the ConditionalQA and HotpotQA datasets.}
\label{table:all_results}
\begin{adjustbox}{width=0.99\textwidth,totalheight=\textheight,keepaspectratio}
\begin{tabular}{|l|l|cccc|ccc|}
\hline
\multirow{2}{*}{\textbf{\begin{tabular}[c]{@{}l@{}}LLM/\\ Model \\ used\end{tabular}}} & \multirow{2}{*}{\textbf{Approach}} & \multicolumn{4}{c|}{\textbf{Dataset 1: ConditionalQA}} & \multicolumn{3}{c|}{\textbf{Dataset 2: HotpotQA}} \\ %\cline{3-9}
 &  & \begin{tabular}[c]{@{}c@{}}\textbf{Avg F1 Score}\\ {[}All {]}\end{tabular} & \begin{tabular}[c]{@{}c@{}}\textbf{F1 Score }\\ {[}Yes/No {]}\end{tabular} & \begin{tabular}[c]{@{}c@{}}\textbf{F1 Score }\\ {[}Span {]}\end{tabular} & \begin{tabular}[c]{@{}c@{}}\textbf{F1 Score } \\ {[}Conditional {]}\end{tabular} & \begin{tabular}[c]{@{}c@{}}\textbf{Avg F1 Score}\\ {[}All {]}\end{tabular} & \begin{tabular}[c]{@{}c@{}}\textbf{F1 Score }\\ {[}Yes/No {]}\end{tabular} & \begin{tabular}[c]{@{}c@{}}\textbf{F1 Score }\\ {[}Span {]}\end{tabular} \\ \hline

\multirow{5}{*}{Mistral} 
& Vanilla LLM & 31.84 & 52.83 & 8.4 & 25.11 & 24.31 & 68.00 & 22.01 \\ %\cline{2-9}
& Code Prompt & 28.26 & 41.74 & 16.30 & 22.76 & 09.69 & 00.00 & 09.69 \\ %\cline{2-9}
& Text Prompt & 27.36 & 31.58 & 25.64 & 23.51 & 16.94 & 04.00 & 17.62 \\ %\cline{2-9}
& Think-on-Graph & 16.24 & 21.58 & 12.05 & 5.53 & 04.76 & 04.00 & 04.80 \\ %\cline{2-9}
& Vanilla RAG & 40.12 & 52.88 & 25.86 & 21.11 & 49.41 & 56.00 & 49.06 \\ %\cline{2-9}
& HybridContextQA & 45.17 & 60.00 & 33.55 & 34.39 & 51.16 & 32.00 & 52.17 \\ %\cline{2-9}
& \textbf{KGCaRe (ours)} & \textbf{57.89} & \textbf{73.42} & \textbf{42.08} & \textbf{57.17} & \textbf{53.53} & \textbf{72.00} & \textbf{54.00} \\ \hline

\multirow{5}{*}{Mixtral} 
& Vanilla LLM & 41.35 & 62.29 & 17.96 & 39.83 & 27.45 & 56.00 & 25.95 \\ %\cline{2-9}
& Code Prompt & 40.88 & 44.80 & \textbf{40.99} & 34.62 & 11.07 & 28.00 & 10.18 \\ %\cline{2-9}
& Text Prompt & 46.60 & 56.95 & 40.15 & 39.36 & 22.35 & 20.00 & 22.47 \\ %\cline{2-9}
& Think-on-Graph & 17.40 & 23.14 & 12.90 & 05.54 & 04.87 & 00.00 & 05.13 \\ %\cline{2-9}
& Vanilla RAG & 53.54 & 73.54 & 30.91 & \textbf{49.46} & 53.84 & 48.00 & 54.15 \\ %\cline{2-9}
& HybridContextQA & 53.71 & 74.13 & 36.45 & 42.59 & 53.68 & 48.00 & 53.98 \\ %\cline{2-9}
& \textbf{KGCaRe (ours)} & \textbf{59.45} & \textbf{78.32} & 38.37 & 47.93 & \textbf{58.27} & \textbf{56.00} & \textbf{58.39} \\ \hline

\multirow{5}{*}{GPT 3.5}
& Vanilla LLM & 33.02 & 61.88 & 24.27 & 35.10 & 38.86 & 76.00 & 36.91 \\ %\cline{2-9}
& Code Prompt & 48.27 & 70.20 & 23.76 & 41.36 & 54.11 & 84.00 & 52.54 \\ %\cline{2-9}
& Text Prompt & 57.15 & \textbf{73.13} & 45.54 & 46.25 & 60.23 & 88.00 & 58.77 \\ %\cline{2-9}
& Think-on-Graph & 16.29 & 13.97 & 20.67 & 09.48 & 14.32 & 16.00 & 14.24 \\ %\cline{2-9}
& Vanilla RAG & 57.79 & 71.91 & 42.01 & 54.27 & 61.84 & \textbf{96.00} & 60.05 \\ %\cline{2-9}
& HybridContextQA & 55.03 & 71.21 & 42.97 & 43.52 & 61.79 & 92.00 & 60.20 \\ %\cline{2-9}
& \textbf{KGCaRe (ours)} & \textbf{60.01} & 72.02 & \textbf{46.59} & \textbf{54.83} & \textbf{71.48} & 88.00 & \textbf{70.61}  \\ \hline

\multirow{5}{*}{GPT 4o} 
& Vanilla LLM & 34.72 & 58.50 & 08.15 & 14.78 & 50.53 & 96.00 & 48.15 \\ %\cline{2-9}
& Code Prompt & 53.29 & 76.92 & 26.89 & 36.36 & 68.82 & 92.00 & 67.60 \\ %\cline{2-9}
& Text Prompt & 59.16 & 81.82 & 40.32 & 48.20 & 65.07 & 84.00 & 64.07 \\ %\cline{2-9}
& Think-on-Graph & 20.40 & 21.45 & 21.46 & 07.85 & 10.79 & 00.00 & 11.36 \\ %\cline{2-9}
& Vanilla RAG & 62.26 & 79.48 & 43.01 & 45.32 & 65.05 & 92.00 & 63.64 \\ %\cline{2-9}
& HybridContextQA & 63.71 & 81.58 & \textbf{50.71} & \textbf{51.99} & 73.37 & 96.00 & 72.18 \\ %\cline{2-9}
& \textbf{KGCaRe (ours)} & \textbf{67.55} & \textbf{83.10} & 50.05 & 48.90 & \textbf{80.21} & \textbf{96.00} & \textbf{79.38} \\ \hline

\end{tabular}
\end{adjustbox}
%\vspace{-1.5em}
\end{table*}

Table~\ref{table:all_results} shows that on both the datasets across all models, the proposed KGCaRe approach consistently achieves the highest performance considering all the questions (Avg F1 Score), validating the effectiveness of the hybrid retrieval and reasoning strategy.

On ConditionalQA, for the Mistral model, baseline methods perform poorly, with Vanilla LLM and Code Prompt achieving an average F1 of 31.84 and 28.26, respectively, and Text Prompt scoring 27.36. The KG-only approach (Think-on-Graph) performs worse at 16.24, reflecting its limitations in better context retrieval from KG and answer generation for the complex conditional questions in comparison to other approaches. Vanilla RAG and HybridContextQA improve performance to 40.12 and 45.17, respectively, while KGCaRe significantly boosts the average F1 to 57.89. Notably, it achieves strong gains in the conditional category (F1 = 57.17) and Yes/No category (F1 = 73.42), underscoring its ability to handle multi-hop and condition-sensitive reasoning even with small models. On HotpotQA, for the Mistral model, KGCaRe outperforms all other existing approaches for all QA categories.

Mixtral, a more capable open-source model, shows higher baseline performance. On ConditionalQA, Vanilla LLM, Code Prompt and Text Prompt achieve 41.35, 40.88 and 46.60 average F1 scores, respectively, with Think-on-Graph again underperforming at 17.40. Vanilla RAG and HybridContextQA perform well with an average F1 of 53.54 and 53.71, respectively. KGCaRe outperforms all baselines with an average F1 of 59.45, showing robust performance. On HotpotQA, for the Mixtral model, KGCaRe again outperforms all other existing approaches for all QA categories. These results highlight the value of hybrid retrieval in moderately strong LLMs, especially when symbolic reasoning is paired with semantic context.

GPT-3.5 delivers strong baseline results, on ConditionalQA, with Vanilla LLM (Avg F1 = 33.02), Code Prompt (Avg F1 = 48.27) and Text Prompt (Avg F1 = 57.15) outperforming the Think-on-Graph method (Avg F1 = 16.29) by a wide margin. Interestingly, Text Prompt yields the highest Yes/No F1 (73.13) across all GPT-3.5 configurations. However, KGCaRe achieves the best overall performance (Avg F1 = 60.01) and the highest conditional F1 (54.83), as well as the best span-based F1 (46.59). On HotpotQA, for the GPT-3.5 model, KGCaRe again outperforms all other existing approaches for all QA categories, except Yes/No. This shows that while prompting alone performs well, complex reasoning benefits from the hybrid design.

As the most powerful model in the comparison, GPT-4o achieves high scores across all methods. On ConditionalQA, Vanilla LLM, Code Prompt and Text Prompt reach average F1 scores of 34.72, 53.29 and 59.16. Think-on-Graph remains ineffective even in this setting (Avg F1 = 20.40). HybridContextQA reaches an average F1 of 63.71, and KGCaRe improves further to 67.55, achieving the highest Yes/No F1 (83.10) and competitive results in the span (50.05) and conditional (48.90) categories. On HotpotQA, for the GPT-4o model, KGCaRe again outperforms all other existing approaches with the highest F1 score in all categories, except the Yes/No category, where HybridContextQA and KGCaRe achieve similar F1 scores. These overall gains confirm the utility of symbolic augmentation even in high-capacity LLMs.

The results in Table~\ref{table:all_results} clearly show that Think-on-Graph lags significantly across all models and question types. On HotpotQA, Think-on-Graph remains ineffective and produces an F1 score of 0 with Mixtral and GPT-4o for the Yes/No category. We found that this occurs because both models generate verbose answers without explicitly including the word ``yes” or ``no”. Vanilla RAG, Code and Text Prompt perform reasonably well on Yes/No and span questions, especially with stronger LLMs. KGCaRe consistently outperforms all baselines across models and question types at Avg F1 Score, and outperforms most models for Yes/No, Span, and Conditional types,  demonstrating the effectiveness of integrating symbolic KG traversal with vector-based semantic retrieval. These results support our core hypothesis: combining structured and unstructured context, and reasoning over both, yields superior performance on complex, conditional QA tasks, regardless of the underlying LLM's capacity. The largest performance margins are observed in conditional and Yes/No questions, where reasoning steps are often multi-hop, conditional, and dispersed across documents, highlighting situations where the hybrid model excels.

\subsection{Ablation Study}
During error analysis on the ConditionalQA dataset, we observed that out of the 136 Yes/No type questions, 30 had multiple valid answers in the ground truth. To better evaluate the performance of KGCaRe and existing approaches, we focused on the subset of Yes/No questions with a single definitive answer. Table~\ref{table:yesno_f1} presents a comparative analysis of KGCaRe and baseline methods on these 106  Yes/No QA pairs having a single definitive answer.

\begin{table}
\centering
\small
\caption{Comparative analysis of KGCaRe and existing approaches on 106 Yes/No QA pairs from the ConditionalQA dataset.}
\label{table:yesno_f1}
\begin{tabular}{|l|l|c|}
\hline
\textbf{LLM / Model used} & \textbf{Approach} & \begin{tabular}[c]{@{}c@{}}\textbf{F1 Score }\\ {[}Yes/No type QA {]}\end{tabular}  \\ \hline

\multirow{5}{*}{Mistral}
& Vanilla LLM & 58.49 \\
& Code Prompt & 40.58 \\
& Text Prompt & 30.89 \\
& Think-on-Graph & 39.94 \\
& Vanilla RAG & 62.26 \\
& HybridContextQA & 57.13 \\
& \textbf{KGCaRe (ours)} & \textbf{69.81} \\ \hline

\multirow{5}{*}{Mixtral}
& Vanilla LLM & 64.22 \\
& Code Prompt & 43.86 \\
& Text Prompt & 57.34 \\
& Think-on-Graph & 45.65 \\
& Vanilla RAG & 74.84 \\
& HybridContextQA & 76.27 \\
& \textbf{KGCaRe (ours)} & \textbf{82.07} \\ \hline

\multirow{5}{*}{GPT 3.5}
& Vanilla LLM & 66.03 \\
& Code Prompt & \textbf{72.07} \\
& Text Prompt & 70.95 \\
& Think-on-Graph & 44.34 \\
& Vanilla RAG & 70.75 \\
& HybridContextQA & 68.86 \\
& \textbf{KGCaRe (ours)} & 71.75 \\ \hline

\multirow{5}{*}{GPT 4o}
& Vanilla LLM & 71.69 \\
& Code Prompt & 84.90 \\
& Text Prompt & 85.85 \\
& Think-on-Graph & 43.03 \\
& Vanilla RAG & 83.96 \\
& HybridContextQA & 80.97 \\
& \textbf{KGCaRe (ours)} & \textbf{88.67} \\ \hline

\end{tabular}
%\vspace{-2.38em}
\end{table}

For the Mistral model, the Vanilla LLM, Code Prompt, and Think-on-Graph baselines achieve modest F1 scores of 58.49, 40.58 and 39.94, respectively (see Table~\ref{table:yesno_f1}), while Text Prompt performs slightly worse at 30.89. The HybridContextQA and Vanilla RAG improve performance significantly to 57.13 and 62.26, respectively, and KGCaRe further enhances it to 69.81, demonstrating the benefit of combining symbolic reasoning with neural retrieval even in smaller open-source LLMs.

With Mixtral, all approaches perform better than on Mistral, showing Mixtral's stronger capability on the QA task. The baseline Vanilla LLM and Text Prompt achieve 64.22 and 57.34, respectively, while Think-on-Graph performs slightly lower at 45.65. Vanilla RAG and HybridContextQA achieve an F1 score of 74.84 and 76.27, respectively, and KGCaRe again improves on this, achieving the highest score of 82.07. This shows that the hybrid reasoning strategy is particularly effective when leveraged with more powerful open-source models.

For GPT-3.5, the Code Prompt surprisingly performs very well with an F1 score of 72.07, surpassing Vanilla LLM (66.03), Text Prompt (70.95), Think-on-Graph (44.34), Vanilla RAG (70.75), and even HybridContextQA (68.86). However, KGCaRe still manages to slightly improve over HybridContextQA, reaching an F1 score of 71.75. Although the gain is smaller compared to open-source models, this suggests that GPT-3.5 also benefits from hybrid context to a limited but still positive extent.

In the case of GPT-4o, both Code Prompt and Text Prompt already perform extremely well with F1 scores of 84.90 and 85.85, respectively. Despite this strong baseline, KGCaRe still achieves the highest F1 score of 88.67, marginally outperforming all other approaches. This indicates that even for high-capacity LLMs, KGCaRe can offer marginal yet meaningful improvements, particularly by enhancing reasoning consistency and contextual faithfulness.

Across all LLMs, we observe that KGCaRe consistently outperforms the KG-only (Think-on-Graph) and vector-only (Text Prompt, Code Prompt, and Vanilla RAG) baselines, reaffirming the value of combining symbolic and neural retrieval methods. The largest relative gains are observed in Mistral and Mixtral, where the base model’s reasoning capabilities benefit most from structured traversal and context integration. In summary, Table~\ref{table:yesno_f1} illustrates that KGCaRe achieves the best or competitive performance across most model settings on Yes/No questions with a single definitive answer. 

\subsection{Explanation and Visualization}

\begin{figure*}[t]
\centering
\includegraphics[width=0.99\textwidth]{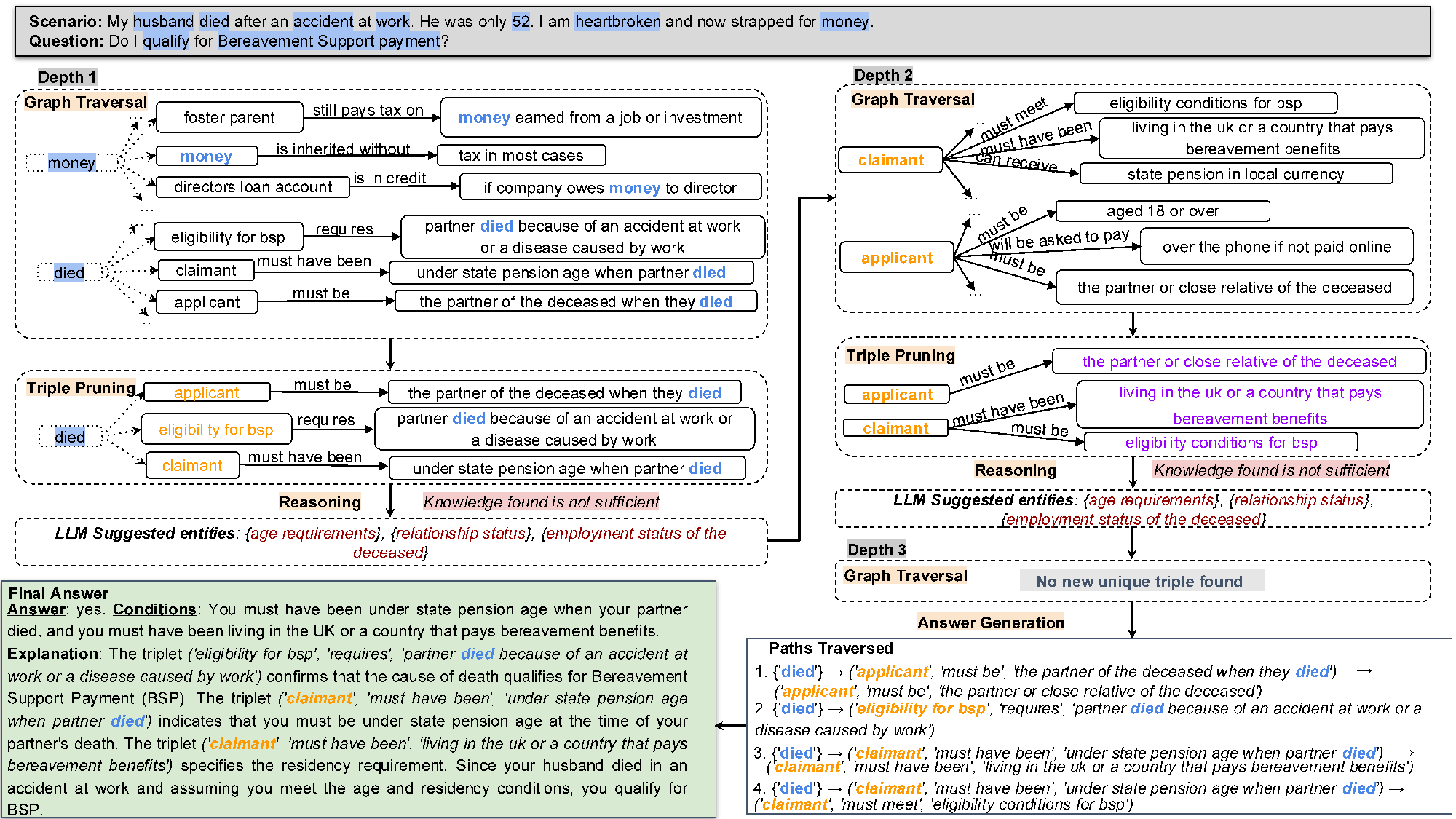} % Reduce the figure size so that it is slightly narrower than the column.
\caption{Explaining the Knowledge Graph traversal steps of the proposed algorithm using an example.}
\label{fig_KG_Paths}
%\vspace{-1.5em}
\end{figure*}

Our proposed approach can also provide an explanation for the generated answer by showing the paths in the KG that lead to that answer (see Figure \ref{fig_KG_Paths}). As shown in Figure \ref{fig_KG_Paths}, we have a complex question as an example from the ConditionalQA dataset. We have a complex and conditional question from an applicant whose husband died at work. The applicant has mentioned some details and wants to know about the eligibility for the Bereavement Support Payment.

Now,  to extract the relevant context from the KG, our KGCaRe approach will go through the KG for a specified number of depths until it extracts the satisfactory context to answer the question or satisfy the termination condition, which is the number of depths. Then it uses the knowledge or paths traversed from all the depths to generate the final answer. In Depth 1, it looks into the KG to get keyword matches of the entities from the question. There are different ways to do this matching, but here it uses the direct keyword match. We can see here matches for the two keywords from the question: ``money” and ``died”. We can also see the associated triples extracted from the KG. Here, initially, it extracts all the information, and then it performs pruning. In pruning, it keeps only those triples that are most relevant for answering the question. In this case, it found the three most relevant triples as shown in Figure \ref{fig_KG_Paths}.

Then it performs reasoning using an LLM. It asks the LLM if the information found so far is enough to answer the question. In the case of Depth 1, the LLM finds that the information is not sufficient. Therefore, the approach asks the LLM to suggest other entities to search for in the KG and which might help to answer the question. Here we can see the LLM suggested entities as shown in Figure \ref{fig_KG_Paths}. It also saves these 3 triples to a variable named ``Path Traversed” so that it has a record of them, and it can use them in answer generation.

Now it goes into the next depth, that is Depth 2, and looks for expanding with keywords from what we saw in the previous depth, like ``claimant”, ``applicant”, etc. It finds some new information related to these, which it will then prune to keep only those relevant to the question (three more triples). Now the approach again asks the LLM if, at this depth, the information is sufficient to answer the question. The LLM still decides this is not enough to answer the question and returns some more clue entities which might be useful. Now, before it goes to the next depth, it again saves the triples found so far in Depth 2 to ``Path Traversed”. It found one new triple related to ``applicant” that it adds to the existing path. It also found two more triples related to the claimant, one states that they must have lived in the UK, and another states that the claimant must be eligible for BSP.

Now it goes to the next depth that is Depth 3, and this time it didn't find any new unique triple, and since this is the last depth considered by our  approach, it generates the answer to the question using the information of paths it has so far.
The final answer generated is Yes, and the condition is that the applicant must be under pension age and must be living in the UK to be eligible for Bereavement support. It also provides an explanation showing which triples were used to reach this answer, as shown in Figure \ref{fig_KG_Paths}. 

Overall, our approach enhances trust by providing human-understandable explanations for the generated answer. 

\section{Conclusion} \label{sec_conclusions}

In this work, we proposed KGCaRe, a hybrid retrieval-augmented question answering approach that tightly integrates symbolic reasoning over a knowledge graph with neural retrieval from a vector store. Our approach is specifically designed to address the challenges of answering complex conditional domain-specific questions—an area where general-purpose LLMs often fall short due to missing context or inability to interpret conditions.

By constructing a KG using a multi-prompt LLM-based pipeline and pairing it with FAISS-based neural retrieval, KGCaRe leverages complementary strengths of both structured and unstructured context. During inference, KGCaRe performs iterative graph traversal guided by an LLM, prunes irrelevant paths, and dynamically updates its memory to support multi-hop reasoning. The final answer is generated using a prompt that combines the curated triples and semantically retrieved text, allowing for both factual accuracy and traceable explainability.

We evaluated KGCaRe on two complex conditional QA datasets, demonstrating consistent improvements across all answer types and LLMs compared to strong baselines such as Vanilla LLM, Text Prompt, Code Prompt, Think-on-Graph, Vanilla RAG, and HybridContextQA. In particular, our KGCaRe approach shows significant gains for conditional answers, validating our hypothesis that hybrid retrieval can better handle multi-faceted reasoning in complex domains.

Looking forward, our framework opens new directions for explainable QA systems in high-stakes settings like healthcare, law, and public policy, where both precision and transparency are critical. \\

\section*{Declarations}

\bmhead{Funding}

This publication has emanated from research conducted with the financial support of Research Ireland under Grant Number 12/RC/2289\_P2 - Insight Research Ireland Centre for Data Analytics and a grant from Fidelity Investments. For the purpose of Open Access, the author has applied a CC BY public copyright licence to any Author Accepted Manuscript version arising from this submission.

\bmhead{Data Availability}
The datasets used in this study are publicly available.

\bmhead{Code availability}
Code is available at the following GitHub repository: \\ 
\url{https://github.com/GhanshyamVerma/KGCaRe}

\bibliography{sn-bibliography}% common bib file
%% if required, the content of .bbl file can be included here once bbl is generated
%%\input sn-article.bbl

\begin{appendices}

\section{Data Appendix}
\label{sec:Data_Appendix}
This section provides further details about the dataset used for experiments.

\subsection{ConditionalQA}

In the ConditionalQA dataset \cite{sun-etal-2022-conditionalqa}, each question-answer pair consists of three components. The first is the document itself, which is a structured policy text organized into hierarchical sections and often contains cross-references to other sections. These documents were scraped from official UK government websites and processed into structured text by serializing the Document Object Model (DOM) trees into a flattened sequence of HTML elements \cite{sun-etal-2022-conditionalqa}. The second component is the question, which is typically centered around eligibility, procedural compliance, or exceptions and may require reasoning across multiple parts of the document. These questions vary in type and include yes/no responses, extractive span-based answers, and cases that may not be answerable given the context. The third component is the user/applicant scenario, which adds contextual background representing a specific user’s situation or constraint. This narrative helps simulate real-world query complexity and often includes implicit conditions that must be resolved during the process of generating answers. 

For the ConditionalQA dataset \cite{sun-etal-2022-conditionalqa}, we use the development set as the evaluation benchmark for all comparative experiments, as the test set is not publicly available. ConditionalQA consists of 2,338 training QA pairs and
285 development QA pairs. Out of 285 questions from the ConditionalQA development set, we removed the fourteen unanswerable questions. After removing the unanswerable questions, in the ConditionalQA development set, we were left with 271 questions, as shown in Table \ref{table:dataset_stats}.  

The ConditionalQA dataset has three types of questions: Yes/No type, Span (extractive) type, and conditional type. The ConditionalQA development set has 143 Yes/No QA pairs, 102 Span QA pairs, and 63 Conditional QA pairs.

\subsection{HotpotQA}
The HotpotQA dataset \cite{yang2018hotpotqa} is a large-scale complex QA dataset that requires reasoning to be performed over multiple documents to generate the answer to the multi-hop questions. This dataset is created using crowdsourcing based on Wikipedia articles, showing multiple context documents to crowd workers such that they can create questions that require multi-hop reasoning.  

HotpotQA mainly has two types of questions: Yes/No and Span (extractive) type. The span type questions include the questions that require bridge entity identification and comparison between two entities to answer the multi-hop questions \cite{yang2018hotpotqa}.  

The HotpotQA train set has 90,564 QA pairs, having a combination of easy, medium, and hard questions. We evaluate the existing and proposed approaches on a subset of the HotpotQA development set. The original HotpotQA development set had 7,405 QA pairs. All the questions in the development set are hard/complex multi-hop questions. We selected a set of 500 QA pairs (see Table \ref{table:dataset_stats}) from the original development set using random stratified sampling. The HotpotQA development set has 25 Yes/No QA pairs and 475 Span QA pairs.

\begin{table}[h]
\centering
\small
\caption{Number of QA pairs in the training and development sets of the ConditionalQA and HotpotQA datasets. }
\label{table:dataset_stats}
\begin{tabular}{|l|c|c|}
\hline
\textbf{Dataset} & \textbf{Train Set} & \textbf{Dev Set} \\ \hline
ConditionalQA & 2,338 & 271  \\
HotpotQA & 90,564 & 500 \\ \hline
\end{tabular}
\end{table}

Datasets and the code to preprocess them are available at the GitHub repository: \\
\url{https://github.com/GhanshyamVerma/KGCaRe}

\section{Knowledge Graph Appendix}
\label{sec:KG_Appendix}
We constructed two KGs using our proposed KGCaRe approach: one from ConditionalQA documents and another from HotpotQA documents. The KG constructed using ConditionalQA has 7,031 triples and 7,366 entities, while the KG constructed using HotpotQA has 17,776 triples and 13,286 entities, as shown in Table \ref{table:KG_stats}.

\begin{table}[h]
\centering
\small
\caption{Number of entities and triples in the generated KGs using KGCaRe.}
\label{table:KG_stats}
\begin{tabular}{|l|cc|cc|}
\hline
\textbf{KG Creation Approach} 
& \multicolumn{2}{c|}{\textbf{ConditionalQA Dataset}}
& \multicolumn{2}{c|}{\textbf{Hotpot QA Dataset}} \\ 
\hline
& \textbf{Entities}
& \textbf{Triples}
& \textbf{Entities}
& \textbf{Triples} \\ 
\hline
KGCaRe
& 7366
& 7031
& 13286
& 17776 \\ 
\hline
\end{tabular}
\end{table}

Knowledge Graphs and the developed software pipeline to construct them are available at the GitHub repository: \\
\url{https://github.com/GhanshyamVerma/KGCaRe}

\section{Code Appendix}
\label{sec:Code_Appendix}
Code for the existing approaches and the proposed KGCaRe approach is available at the following GitHub repository: \\ 
\url{https://github.com/GhanshyamVerma/KGCaRe}

\section{Technical Appendix}
\label{sec:Technical_Appendix}
This section provides technical details.

\subsection{Evaluation} \label{sec:Evaluation_Appendix}
\textbf{Avg F1 score: } The Avg F1 score represents the calculated average F1 score using the exact match policy between predicted and ground truth answer, considering all the questions of the development set \cite{sun-etal-2022-conditionalqa, yang2018hotpotqa}.

\textbf{F1 score: } The F1 score represents the calculated F1 score using the exact match policy between the predicted and ground truth answer \cite{sun-etal-2022-conditionalqa, yang2018hotpotqa}. 

The evaluation scripts were provided with the publicly available benchmark datasets, ConditionalQA \cite{sun-etal-2022-conditionalqa} and HotpotQA \cite{yang2018hotpotqa}. We used the same scripts for our evaluation. Our code repository includes the evaluation script as well. Link to the code: \url{https://github.com/GhanshyamVerma/KGCaRe}

\subsection{Implementation Details} \label{sec:Implementation_Appendix}

\begin{itemize}

    \item \textbf{Code Prompt}:
    \begin{description}
        \item[Note:] Same parameters were used as mentioned in \cite{puerto2024code}.
    \end{description}
    
    \item \textbf{Text Prompt}:
    \begin{description}
        \item[Note:] Same parameters were used as mentioned in \cite{puerto2024code}. 
    \end{description}
    
    \item \textbf{Think on Graph}:
    \begin{description}
        \item[Temperature:] 0.1
        \item[Maximum Output Length:] 256
        \item[Width and Depth of Exploration:] 3
        \item[Note:] Temperature is set to 0.1 for all models. The rest of the parameters are the same as mentioned in \cite{sun2023think}.
    \end{description}

    \item \textbf{HybridContextQA}:
    \begin{description}
        \item[Note:] Same parameters were used as mentioned in \cite{verma2024hybridcontextqa}.
    \end{description}
    
    \item \textbf{KGCaRe}:
    \begin{description}
        \item[Experimental Setup:] A combination of commercial APIs and local model inference was used. \\
        
        \item[LLM Models and Prompting: ] \mbox{}\\
        \begin{itemize}
            \item \textbf{GPT-3.5 / GPT-4o} (via OpenAI API using LlamaIndex): Used as-is without modifying default decoding parameters. The default temperature of 0.7 from LlamaIndex was retained.
            \item \textbf{Mistral-7B and Mixtral-8x7B} (via vLLM + OpenAILike interface): Default decoding parameters (e.g., temperature, top\_p) as defined by vLLM were used without manual modification. \\
        \end{itemize}
        
        \item[In-Context Examples:] \mbox{}\\
        \begin{itemize}
            \item \textbf{ConditionalQA:} 4 examples randomly selected from the training set.
            \item \textbf{HotpotQA:} 4 examples randomly selected from the training set. \\
        \end{itemize}
        
        \item[Retrievers:] \mbox{}\\
        \begin{itemize}
            \item \textbf{Vector-based Retriever} (VectorIndexRetriever from LlamaIndex): similarity\_top\_k = 10
            \item \textbf{Knowledge Graph Traversal Retriever:} max\_depth = 3, max\_entities = 20
        \end{itemize}
        These parameters were fixed based on preliminary trials and prior work; no hyperparameter search was conducted. \\
        
        \item[Reranking:] \mbox{}\\
        \begin{itemize}
            \item \textbf{Cohere Reranker:} rerank-english-v3.0 with top\_n = 2
            \item No other configuration parameters were modified.
        \end{itemize}
        
        \item[Selection Criteria:] 
        Most parameters were adopted from the default settings of the respective tools (LlamaIndex, vLLM, Cohere). Deviations, such as reducing the number of in-context examples, were based on practical constraints (e.g., token limits). No grid search or extensive hyperparameter tuning was conducted.
    \end{description}

\end{itemize}

\subsection{Experimental Environment} \label{sec:Experimental_Appendix}

All experiments were conducted on a high-performance Linux machine with the following specifications:

\begin{itemize}
    \item \textbf{Operating System:} Ubuntu 22.04.4 LTS (Jammy)
    
    \item \textbf{CPU:} AMD EPYC 7313P 16-Core Processor
    \begin{itemize}
        \item 32 threads (16 cores, 2 threads per core)
        \item Base frequency: 1.5 GHz
        \item Max frequency: 3.0 GHz
    \end{itemize}
    
    \item \textbf{RAM:} 256 GB
    
    \item \textbf{GPU:} 4 \(\times\) NVIDIA A40 (48 GB each)
    \begin{itemize}
        \item Experiments utilized approximately 43 GB per GPU for inference
    \end{itemize}
    
    \item \textbf{CUDA:}
    \begin{itemize}
        \item CUDA Version: 12.8
        \item CUDA Toolkit: 12.6 (Build V12.6.85)
        \item Driver Version: 570.133.07
    \end{itemize}
    
    \item \textbf{Python Version:} 3.10.16
\end{itemize}

\end{appendices}

\end{document}